\pdfoutput=1

\documentclass[11pt]{article}

\usepackage[]{acl}

\usepackage{subcaption}
\usepackage[T1]{fontenc}
\usepackage{stfloats}
\usepackage{comment}

\usepackage[utf8]{inputenc}

\usepackage{microtype}

\usepackage{inconsolata}
\usepackage{enumitem}
\usepackage{times}
\usepackage{latexsym}
\usepackage{graphicx}
\usepackage{booktabs}
\usepackage{graphicx} 
\usepackage{multirow}
\usepackage{float} 
\usepackage{url}
\usepackage{tcolorbox}
\usepackage{amssymb}

\title{Dialog Flow Induction for Constrainable LLM-Based Chatbots}

\author{Stuti Agrawal\thanks{denotes equal contribution}, Nishi Uppuluri\footnotemark[1], Pranav Pillai\footnotemark[1], Revanth Gangi Reddy, Zoey Li,\\\textbf{Gokhan Tur, Dilek Hakkani-Tur, Heng Ji}\\
University of Illinois Urbana-Champaign \\
\texttt{\{stutia3, nu4, ppillai3, revanth3, shal2\}@illinois.edu}\\
\texttt{\{gokhan, dilek, jih\}@illinois.edu}}

\begin{document}
\maketitle
\begin{abstract}
LLM-driven dialog systems are used in a diverse set of applications, ranging from healthcare to customer service. However, given their generalization capability, it is difficult to ensure that these chatbots stay within the boundaries of the specialized domains, potentially resulting in inaccurate information and irrelevant responses. This paper introduces an unsupervised approach for automatically inducing domain-specific dialog flows that can be used to constrain LLM-based chatbots. We introduce two variants of dialog flow based on the availability of in-domain conversation instances. Through human and automatic evaluation over various dialog domains, we demonstrate that our high-quality data-guided dialog flows\footnote{Code is available at \url{https://github.com/gangiswag/dialog-flows}} achieve better domain coverage, thereby overcoming the need for extensive manual crafting of such flows.
\end{abstract}

\section{Introduction}

The widespread use of Large Language Models (LLMs)~\cite{openai2023gpt4} for chatbots, highlighted by their human-like conversational abilities across many topics, faces challenges in specialized domains due to their tendency to go off-topic. This generalization capability, while a strength, necessitates the development of more effective control mechanisms to ensure chatbots remain within the desired domain of conversation, especially in specialized fields such as healthcare or legal advice.
Controlling LLM-based chatbots can be effectively managed through dialog flows or schemas\footnote{We use the terms \textit{flows} and \textit{schemas} interchangeably. Our definition of dialog schemas follows~\citet{Mosig2020STARAS} to be analogous to task specifications, different from task slots.}~\cite{bohus2009ravenclaw, Mosig2020STARAS}, which structure conversations along predefined paths of dialog actions, acting as directed graphs where nodes represent actions by the user or bot, and edges are the transitions between actions. This structure helps steer the conversation, keeping it within relevant topics, and also enables chatbots to adapt to new tasks or domains without prior training~\cite{zhao-etal-2023-anytod}.


\begin{figure}[t]
    \centering
    \includegraphics[width=1.0\columnwidth]{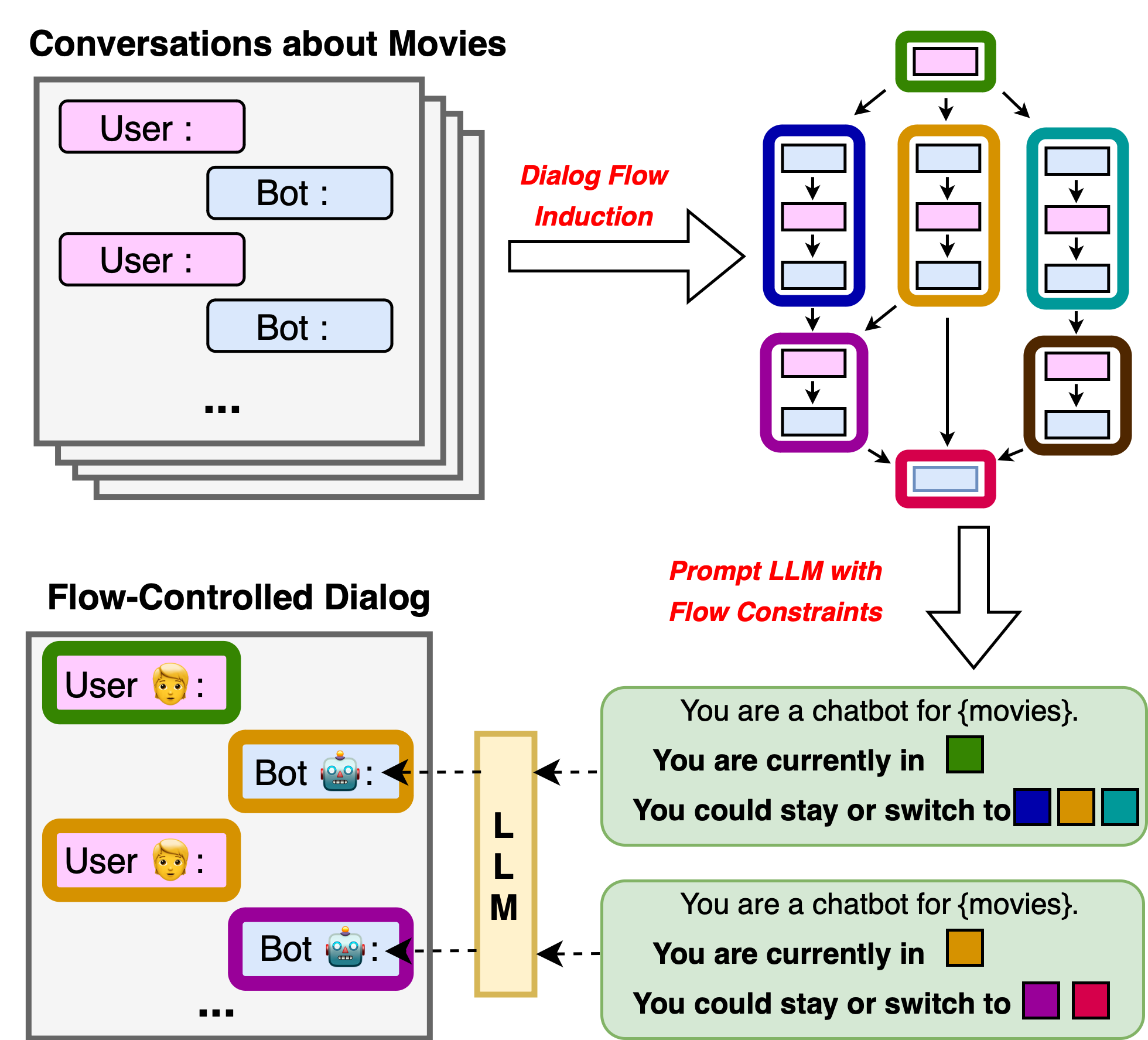}
    \caption{Figure demonstrating how automatically induced domain-specific dialog flows can be used to constrain chatbots to produce domain-focused responses.}
    \label{fig:overall_fig}
    \vspace{-1em}
\end{figure}


 However, the construction of precise dialog flows is challenging~\cite{huang2020challenges}, given the diversity of dialog in different domains. The most prevalent approaches~\cite{mehri2021schema, zhao-etal-2023-anytod} use schemas that are carefully handcrafted by the dialog system developers. The design of dialog schemas thus has significant manual overhead for developers, resulting in scalability and coverage limitations~\cite{zhang2020recent}.

This paper introduces an unsupervised method to generate domain-specific dialog flows, exploiting GPT-4's knowledge to systematically create detailed dialog flows reflecting conversational patterns in various domains. We begin by prompting GPT-4 to produce a structured representation of dialog interactions between users and bots, and then further refine this through self-reflective feedback based on a set of predefined criteria (see figure \ref{fig:llm_flow}).

Further, when we have domain-specific conversations, our approach automatically identifies distinct user and bot dialog actions within these conversations (see figure \ref{fig:data_driven_flow}). These dialog actions, along with selected conversations that exemplify each action, are used to condition the GPT-4 prompt to ensure the dialog flows are grounded using actual domain instances.
This approach enables the automated creation of structured dialog flows, facilitating the development of effective domain-specific chatbots that adhere to their domain's conversational boundaries.
Our main contributions are:
\begin{itemize}[noitemsep]
    \item This paper introduces an approach for automatically constructing dialog flows for various domains in an unsupervised manner. 
    \item The proposed method uses a multi-step framework, that can further leverage domain-specific dialog instances, leading to a graph-like flow illustrating the structure of conversations in the domain.
\end{itemize}

\section{Dialog Flow Induction}

\begin{figure*}[t] 
    \centering
    \includegraphics[width=1.0\textwidth]{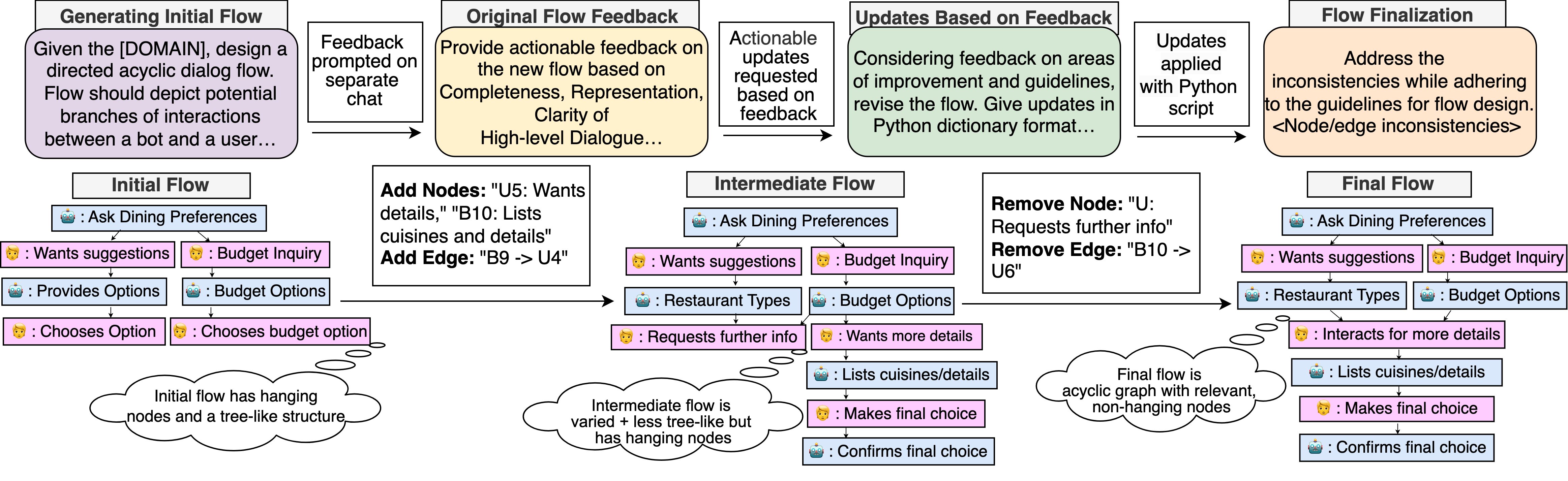}
    \vspace{-0.5em}
    \caption{Figure showing the process for intrinsic flow induction. An initial flow is first generation which is further refined with feedback, update, and clean-up stages. Detailed prompts for each stage are provided in the appendix.}
    \label{fig:llm_flow}
    \vspace{-1em}
\end{figure*}

A dialog flow is a flowchart comprising nodes which can be a user or bot dialog action, and edges that denote logical flow or transitions between these actions. Dialog flows are tailored to different domains.
 Figure \ref{fig:llm_flow} shows an excerpt of a dialog flow, with more detailed examples in the appendix. In this section, we detail our approach for automatically inducing the dialog flow for a given conversation domain. Specifically, we induce two variants of dialog flows, namely \textit{intrinsic flows} (in \S{\ref{sec:intrinsic_flow}}) or \textit{data-guided flows} (in \S{\ref{sec:data_driven_flow}}) depending on whether sample conversations in the domain are available. 

\subsection{Intrinsic Dialog Flow}
\label{sec:intrinsic_flow}
When domain-specific conversation data is unavailable, we propose to induce dialog flows using the \textit{intrinsic} domain-related knowledge of LLMs and their understanding of conversational principles.
Our intrinsic flow induction process starts with GPT-4 creating an initial flow based on the domain's name. Next, GPT-4 self-evaluates the flow based on predetermined guidelines, to provide concrete actionable feedback for improvement. Using this feedback, GPT-4 then suggests a set of edits, which are automatically applied to the initial flow. Finally, automated checks are run to identify inconsistencies in the flow, which GPT-4 then handles in the end clean-up stage. Figure \ref{fig:llm_flow} shows the overall intrinsic flow induction process, with more details on each step provided below.

 \begin{figure*}[t] 
    \centering
    \includegraphics[width=1.0\textwidth]{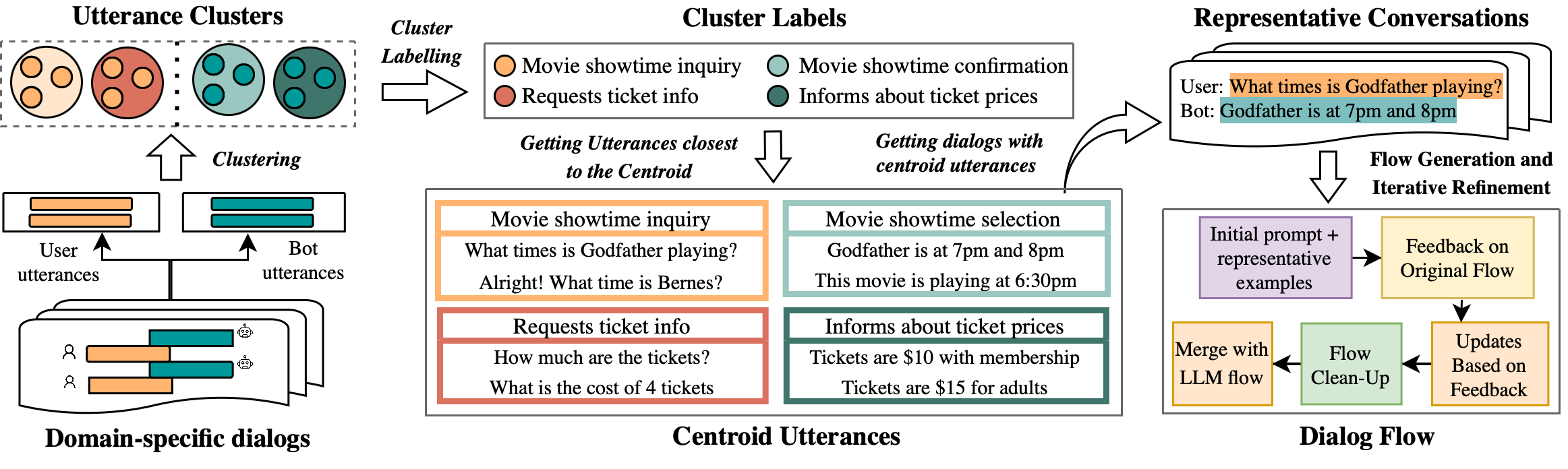}
    \vspace{-0.75em}
    \caption{Figure showing the methodology for inducing dialog flows using a data-guided approach. Representative examples from the domain conversation instances are used to condition the GPT-4 prompts.}    
    \label{fig:data_driven_flow}
    \vspace{-1em}
\end{figure*}



\paragraph{Initial Flow Generation:}
The flow induction starts with prompting GPT-4 with a specific generation prompt to create a dialog flow, as shown in Figure~\ref{fig:llm_flow}. Along with the domain name, the prompt includes details on the intended structure of the dialog flow. After the initial flow is generated, it undergoes further refinement as detailed next.

\paragraph{Flow Feedback and Updates:}
The initial flow often suffers from low coverage along with ambiguous or repetitive action labels for bot and user nodes. We address these by leveraging GPT-4 for self-assessment~\cite{bai2022constitutional} and refining the dialog flow based on the feedback. 
The refinement process starts by obtaining GPT-4 feedback based on the following aspects:
\begin{itemize}[noitemsep]
    \item \textbf{Representativeness:} Both the bot and user actions should be relevant to the domain, and should not be vague or generic.
    \item \textbf{Coverage:} Ensuring the flow captures a broad range of conversational possibilities relevant to the domain.
    \item \textbf{Clarity of Dialog Action:} Each node should reflect a clear and meaningful dialog action. 
    \item \textbf{Optimality:} Eliminate redundancy, ensuring no nodes depict overlapping dialog actions.
\end{itemize}

Based on the shortcomings identified by the self-reflective feedback, GPT-4 is then prompted to output a set of concrete updates to be made to the flow, which can include nodes or edges to add, remove, or edit. To control for the extent to which the flow changes, the updates are performed with an automated Python script rather than directly prompting GPT-4 to apply the updates\footnote{We hypothesize that this provides the ability to heuristically control different aspects of the dialog flow, such as depth, breath, density of edges, etc.}.



\paragraph{Flow Finalization:} Finally, the dialog flow undergoes a clean-up stage where trivial inconsistencies, such as dangling non-terminal nodes, bot-bot or user-user connections, are identified. These are passed as input to GPT-4 along with a final prompt, to ensure the flow is structurally correct.


\subsection{Data-Guided Dialog Flow}
\label{sec:data_driven_flow}

The intrinsic dialog flow induction approach, while expansive in its scope, relies predominantly on the model's inherent knowledge of the typical interactions and transitions that could occur within the specified conversation domain. However, when dialog instances within the given domain are provided, the intrinsic flow can be updated to include actual conversational patterns. We call this approach \textit{data-guided} flow induction, which aims to mirror real-world dialog dynamics.
Specifically, the approach conditions the GPT-4 flow generation prompt with representative examples in the form of action labels and sample conversations for the domain, which help ground the flow to real-life conversation data. Figure \ref{fig:data_driven_flow} gives an overview of data-guided flow induction process, with more details provided below.




\paragraph{Identifying Representative Examples:}
\label{sec:representative_examples}
Given dialog instances for a domain, the following steps identify the user and bot actions, along with sample conversations that are representative of the domain. 
\begin{itemize}[noitemsep]
    \item \textbf{\textit{Clustering and Labeling}}: 
    The user and bot utterances from dialogs in the domain are clustered separately using SentenceBert~\cite{reimers-2019-sentence-bert} embeddings. Next, GPT-4 is prompted to label each cluster with a dialog action by providing it with the utterances closest to each centroid.
    \item \textbf{\textit{Cluster Merging}}: 
    Next, we merge clusters that exhibit significant overlaps in terms of action intent, based on the cosine similarity between the labels. This reduces the redundancy in the action labels by grouping clusters with similar actions.

    \item \textbf{\textit{Picking sample conversations}}: Finally, the conversations that include utterances corresponding to the cluster centroids are picked as the representative dialog instances to include in the GPT-4 prompt for flow generation. This ensures that the conversations encompass a wide spectrum of dialog actions and user intents specific to the domain.

    
\end{itemize}
\paragraph{Flow Generation:}
\label{sec:flow_generation}
As shown in Figure \ref{fig:data_driven_flow}, the flow induction follows a similar generation process as the intrinsic dialog flow. Firstly, the representative action labels and sample conversations for the domain are included in the initial flow generation prompt. Next, the feedback, update, and clean-up steps are applied to result in a dialog flow.  



\paragraph{Merging with Intrinsic Flow:}
The intrinsic flow approach creates broad, expansive dialog flows, but can still fall short of reflecting domain-specific patterns from real-world conversations. On the other hand, solely relying on the domain dialog instances can hurt extensiveness, as they can have limited variability. Hence, we adopt a hybrid approach for the data-guided flow by merging the intrinsic flow with the flow induced solely from domain-specific data. This capitalizes on the extensive scope of the intrinsic flow with the detailed focus from domain data. This merging step is achieved by prompting GPT-4 to identify and retain distinctive features from the intrinsic flow, while removing redundant elements. We call this final flow, \textit{data-guided} flow.

\section{Experiments}
We perform both human and automatic evaluations to assess the induced dialog flows. 

\subsection{Datasets}

Open-domain dialog can involve a single conversation touching upon different domains, such as movies, sports, music, etc. Hence, for simplicity, we consider domains from task-oriented dialog in our experimental settings, wherein the domains are distinct and correspond to the end user task, such as movie tickets, flight booking, restaurant reservations, etc. We consider a dialogs across various task-oriented domains, comprising 24 domains\footnote{We excluded domains that had ambiguous or generic names, such as Play Times, Catalogue, Agreement Bot, etc.} from MetaLWoz~\cite{shalyminov2019few} and 5 domains from MultiWOZ~\cite{budzianowski2018multiwoz}. For the data-guided flow induction, for each domain, we utilized 80\% of the data as domain-specific instances available for training, with the remaining 20\% reserved for evaluating coverage of the bot-bot transitions (described later in \S{\ref{sec:automatic_eval}}).

\subsection{Human Evaluation of Flow Quality}
\label{sec:human_eval}

The evaluators (five undergraduate computer science students) were tasked with examining data-guided and intrinsic flows across the 24 different domains from MetaLwoz. 
The evaluators were given detailed guidelines (provided in the appendix), and were instructed to assess each flow on a scale of 1 to 5 for \textit{domain coverage}, \textit{conclusiveness} and \textit{coherence}.

Table \ref{tab:human_comparison} shows numbers from human evaluation of the data-driven and intrinsic dialog flows. The numbers (expanded to a scale of 20-100) are averaged over all the domains, with flows for each domain being annotated by 5 evaluators. We can see that the data-driven flow, on account of leveraging domain-specific dialog instances, improves over the intrinsic flow on domain coverage. Further, both dialog flows have similarly high scores for conclusiveness and coherence, implying our unsupervised approach, by leveraging GPT-4, can automatically induce high-quality dialog flows. We employed Randolph’s kappa to evaluate the multi-rater agreement. Our findings revealed a kappa value of 0.32, indicating a fair level of agreement across the board. Specifically, the domain coverage metric exhibited the highest kappa value of 0.46, signifying moderate agreement.



\begin{table}[t]
\centering
\begin{tabular}{lccc}
\toprule
 & \textbf{Intrinsic} & \textbf{Data-driven}\\
\midrule
Domain Coverage & 90.7 & \textbf{93.0}  \\
Conclusiveness & \textbf{87.8} & 87.7 \\
Coherence & 84.5 & \textbf{84.8} \\
\bottomrule
\end{tabular}
\caption{Results from human evaluation (in \%) of different aspects of the induced dialog flows}
\label{tab:human_comparison}
\end{table}

\begin{table}[t]
\centering
\begin{tabular}{lcc}
\toprule
\textbf{Dataset} & \textbf{Intrinsic} & \textbf{Data-driven} \\
\midrule
MetaLWoz & 31.6 & \textbf{33.1}\\
MultiWOZ & 39.9 & \textbf{43.0} \\
\bottomrule
\end{tabular}
\caption{Bot-Bot transition coverage (in \%) for the proposed variants of dialog flows on the MetalWoz~\cite{shalyminov2019few} and MultiWOZ~\cite{budzianowski2018multiwoz} datasets. Detailed domain-wise numbers are provided in Table \ref{tab:domain_auto_coverage} in the appendix.}
\label{tab:auto_coverage}
\end{table}

\subsection{Automatic Evaluation of Flow Coverage}
\label{sec:automatic_eval}
Next, we automatically evaluated the domain coverage of different dialog flows, by measuring the coverage on capturing bot-to-bot transitions within the domain conversations in the test set. We leveraged Mistral-7B-Instruct~\cite{jiang2023mistral} to classify bot utterances into the most appropriate node in the dialog flow. We then examined whether the next bot utterance mapped to the directly succeeding node in the dialog flow. Essentially, this metric measures the percentage of bot-bot transitions in domain conversations that conform to the given dialog flow. Table \ref{tab:auto_coverage} shows numbers for automatic coverage evaluation. We can see that the data-driven dialog flow has better coverage of the domain's bot-bot transitions.

\section{Conclusion and Future Work}
We introduce a novel method for developing dialog flows that reflect the combined intrinsic knowledge of LLMs and existing domain-relevant dialogs. Our data-driven dialog flow approach achieves better domain coverage than the intrinsic flow approach across human and automatic evaluations. Our paper outlines a blueprint (in Figure \ref{fig:overall_fig}) for integrating the generated dialog flows into LLM-based chatbots, with a primary focus on the methodologies for dialog flow generation. We believe these dialog flows can be a springboard for future interactive dialog systems that maintain a natural conversation flow within the domain.


%
\section*{Limitations}
In this study, our experimentation was confined to task-oriented dialogs, encompassing a relatively narrow spectrum of dialog flows. This specialization may limit the applicability of our findings to dialog domains characterized by a broader array of tasks and more open-ended dialogues. 
Additionally, our methodology relies solely on unsupervised clustering techniques, bypassing datasets that are annotated with slot values and user intents, which could potentially enhance dialog flow induction. Furthermore, we have not extended our research to test the performance of chatbots constrained by the dialog schemas we developed. Therefore, the efficacy of these schemas in practical chatbot applications remains an area for future investigation.

\section*{Acknowledgment}

We would like to thank the CS STARS program at UIUC for supporting Stuti and Nishi. We are grateful to members of the BlenderNLP group for their valuable comments and feedback. This research is based on
work supported by U.S. DARPA KAIROS Program
No. FA8750-19-2-1004 and U.S. DARPA INCAS Program No. HR001121C0165. The views and conclusions contained
herein are those of the authors and should not be
interpreted as necessarily representing the official
policies, either expressed or implied, of DARPA,
or the U.S. Government. The U.S. Government is
authorized to reproduce and distribute reprints for
governmental purposes notwithstanding any copyright annotation therein.

\bibliography{custom}

\appendix

\clearpage

\section{Appendix}

\begin{table}[!htb]
\centering
\begin{tabular}{lcc}
\toprule
\textbf{MetaLWoz} & \textbf{Intrinsic} & \textbf{Data-driven} \\
\midrule
Alarm set & 32.9 & \textbf{42.2} \\
Apartment finder & 30.9 & \textbf{45.2} \\
Bank bot & \textbf{34.2} & 30.8 \\
Bus schedule & \textbf{37.2} & 14.4 \\
City info & 29.4 & \textbf{33.4} \\
Edit playlist & \textbf{44.2} & 39.4 \\
Event reserve & 28.8 & \textbf{30.5} \\
Library Request & \textbf{35.7} & 30.1 \\
Movie listings & 30.7 & \textbf{34.4} \\
Music suggester & \textbf{34.0} & 25.3 \\
Name suggester & \textbf{43.2} & 16.7 \\
Order pizza & 31.6 & \textbf{36.1} \\
Pet advice & \textbf{33.8} & 31.7 \\
Phone plan & 31.6 & \textbf{37.8} \\
Restaurant picker & \textbf{29.4} & 29.2 \\
Scam lookup & 22.6 & \textbf{31.2} \\
Shopping & 17.0 & \textbf{22.9} \\
Ski Bot & 27.2 & \textbf{32.2} \\
Sports info & 36.6 & \textbf{37.1} \\
Store details & \textbf{35.7} & 32.4 \\
Update calendar & \textbf{38.4} & 28.8 \\
Update contact & \textbf{32.5} & 30.3 \\
Weather check & \textbf{36.1} & 29.5 \\
Wedding planner & 17.0 & \textbf{24.2} \\
\midrule
\textbf{Average} & 31.6 & \textbf{33.1} \\
\bottomrule
\end{tabular}

\begin{tabular}{lcc}
\toprule
\textbf{MultiWOZ} & \textbf{Intrinsic} & \textbf{Data-driven} \\
\midrule
Restaurant & 31.0 & \textbf{43.9} \\
Hotel&43.2&43.2\\
Attractions	&43.3	&\textbf{53.3}\\
Taxi&\textbf{75.3}&50.5 \\
Train&6.9&\textbf{24.1}\\
\midrule
\textbf{Average} & 39.9 & \textbf{43.0} \\
\bottomrule
\end{tabular}
\caption{Bot-Bot transition coverage (in \%) for the proposed variants of dialog flows when measured on various domains in the MetalWoz~\cite{shalyminov2019few} and MultiWOZ~\cite{budzianowski2018multiwoz} datasets.}
\label{tab:domain_auto_coverage}
\end{table}

\begin{table}[t]
\centering
\begin{tabular}{lcc}
\toprule
\textbf{MetaLWoz} & \textbf{Train} & \textbf{Test} \\
\midrule
Alarm set & 1345 & 336 \\
Apartment finder & 399 & 100 \\
Bank bot & 294 & 73 \\
Bus schedule & 718 & 180\\
City info & 772 & 193 \\
Edit playlist & 459 & 115 \\
Event reserve & 431 & 108 \\
Library request & 1071 & 268 \\
Movie listings & 486 & 121 \\
Music suggester & 356 & 89 \\
Name suggester & 399 & 100 \\
Order pizza & 462 & 115 \\
Pet advice & 341 & 85 \\
Phone plan & 397 & 99 \\
Restaurant picker & 428 & 107 \\
Scam lookup & 1326 & 332 \\
Shopping & 722 & 181 \\
Ski bot & 486 & 121 \\
Sports info & 449 & 112 \\
Store details & 590 & 147 \\
Update calendar & 1593 & 398 \\
Update contact & 522 & 131 \\
Weather check & 441 & 110 \\
Wedding planner & 408 & 102 \\
\bottomrule
\end{tabular}
\caption{Statistics of dialogs in various domains in the MetalWoz~\cite{shalyminov2019few} dataset.}
\label{tab:metalwoz_stats}
\end{table}

\begin{figure*}[]
\centering
\includegraphics[width=\textwidth]{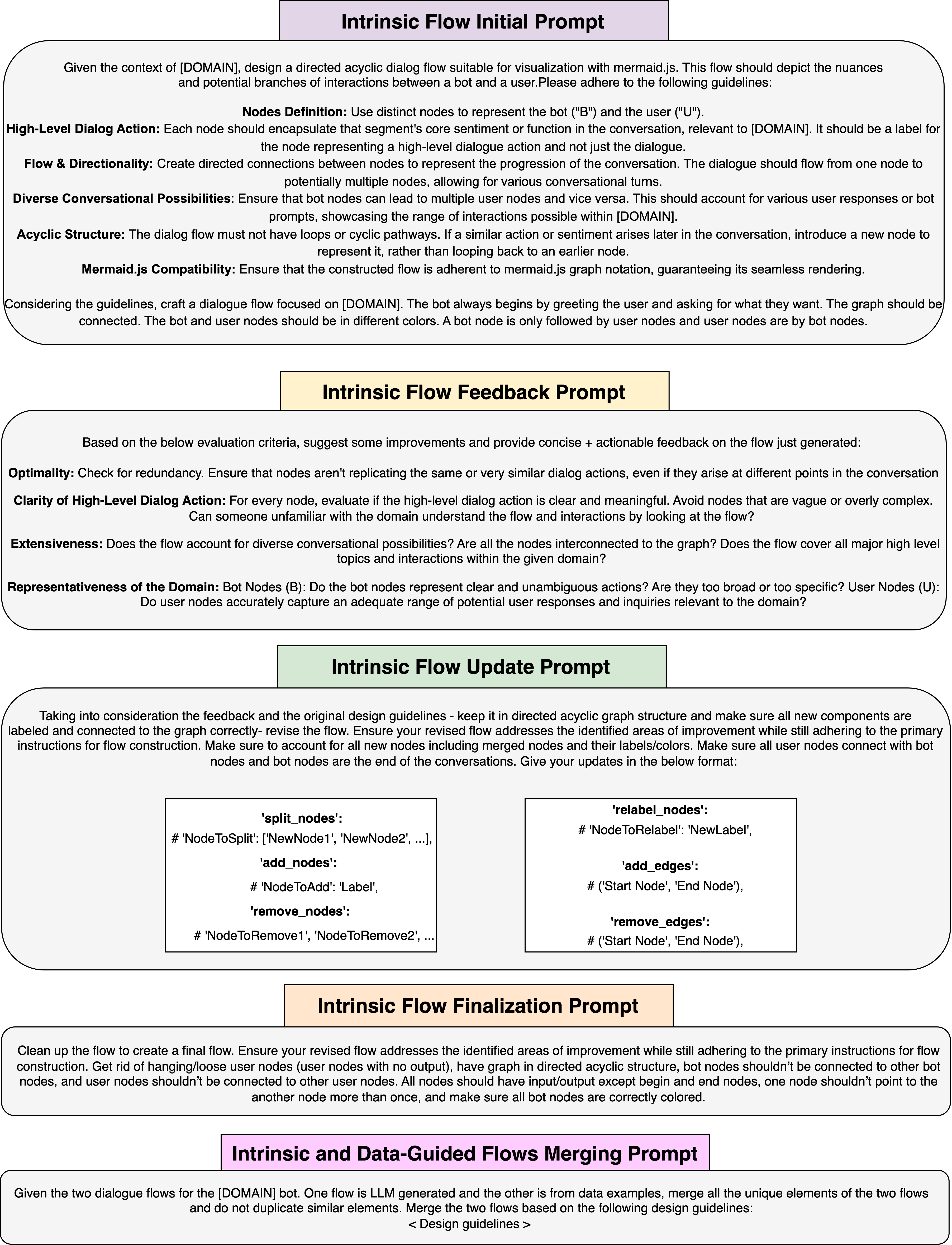}

\captionsetup{width=\textwidth}

\caption{Figure showing prompts for intrinsic and data-guided dialog flow generation.}

\end{figure*}


\begin{figure*}[h]
      \begin{subfigure}[c]{1.0\linewidth}
    \centering
     \includegraphics[width=1\linewidth]{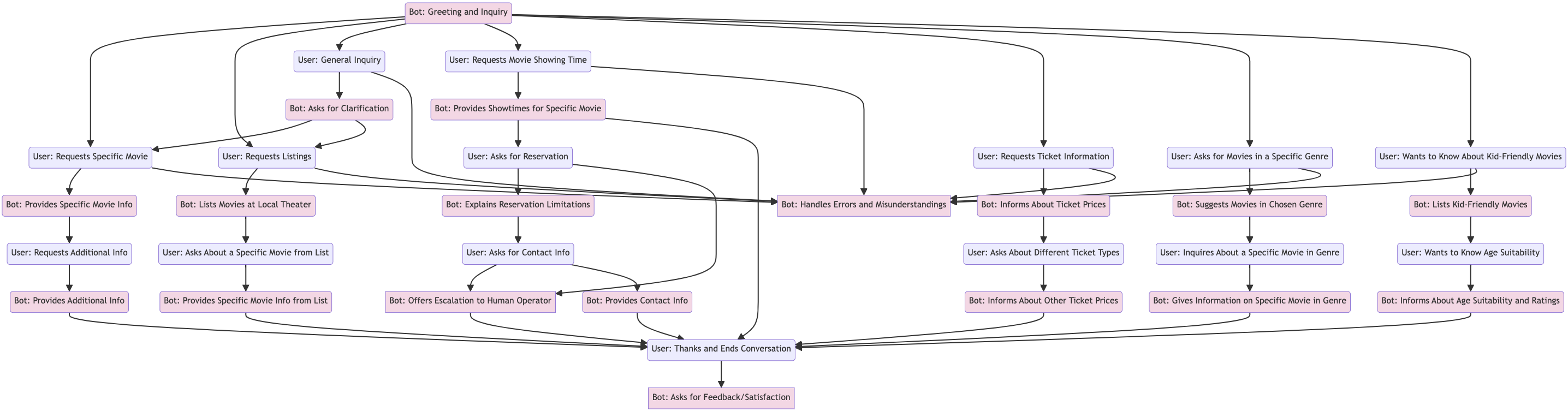}
     \caption{}   
\end{subfigure}
    \hfill
    \begin{subfigure}[c]{1.0\linewidth}
    \centering
     \includegraphics[width=0.6\textwidth]{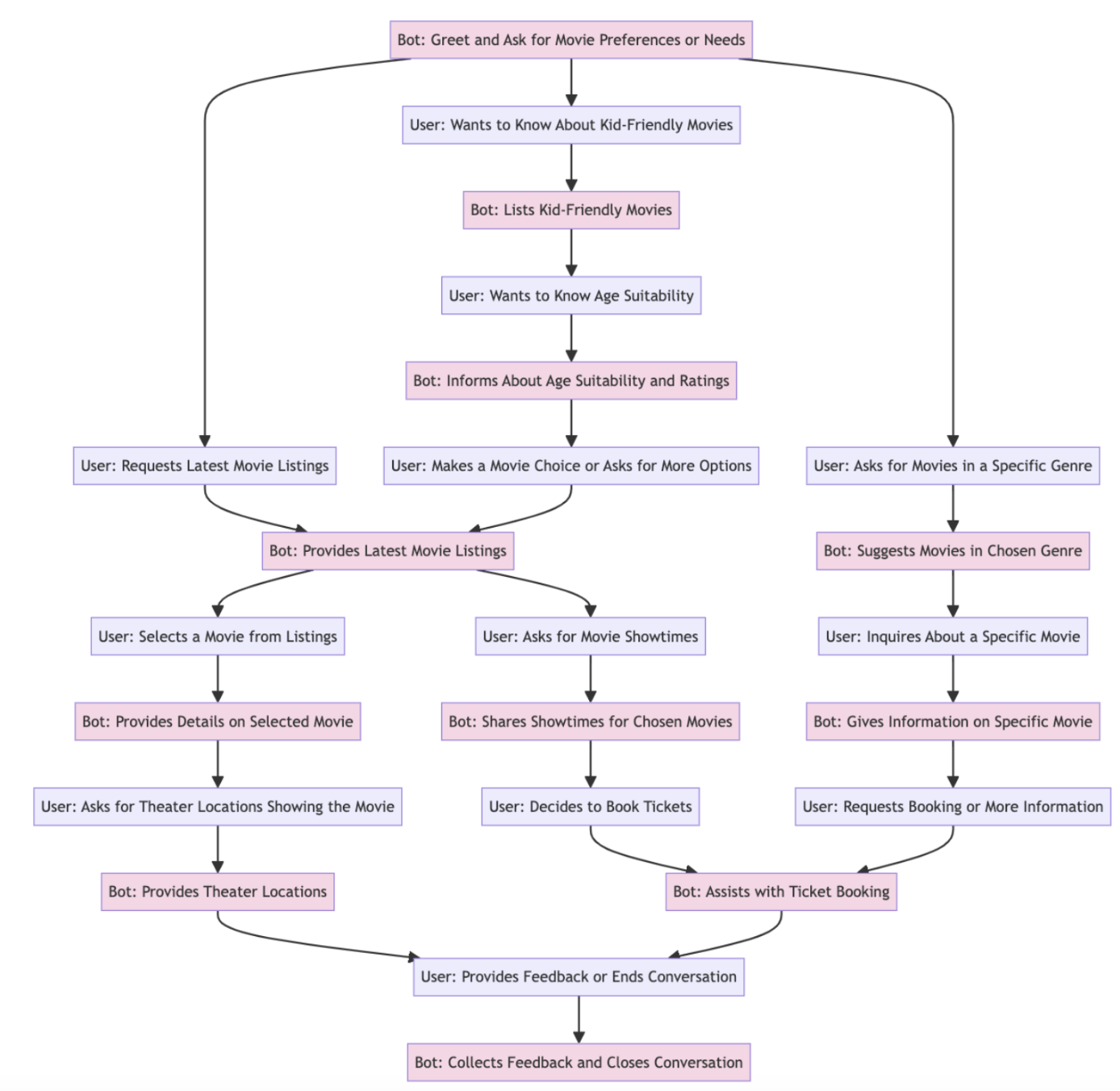}
     \caption{}
\end{subfigure}
\caption{Data-driven (a) and Intrinsic (b) flows for the movie listings domain from MetaLWoz.}
\end{figure*}

\begin{figure*}[h]
      \begin{subfigure}[c]{1.0\linewidth}
    \centering
     \includegraphics[width=1\linewidth]{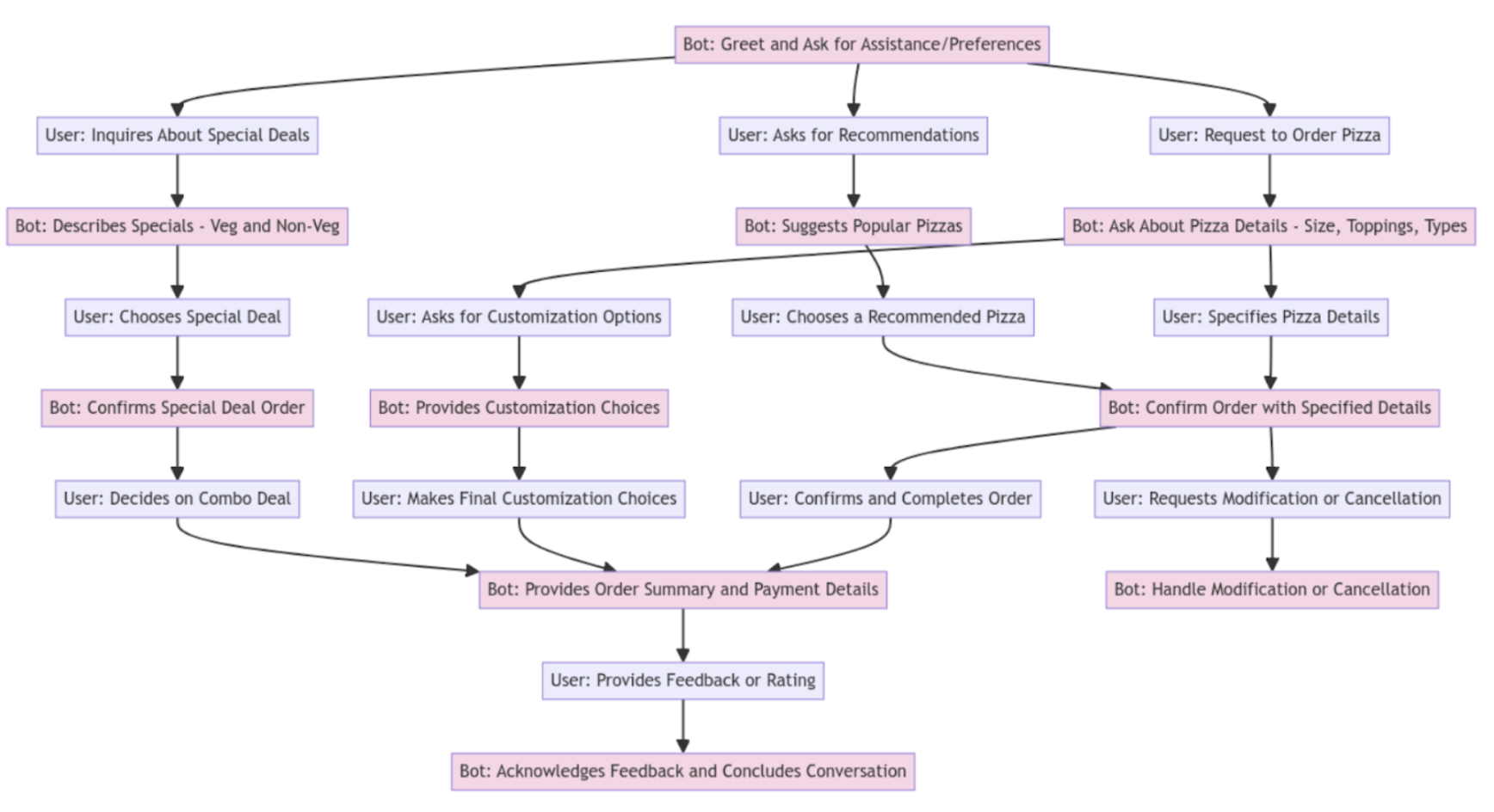}
     \caption{}   
\end{subfigure}
    \hfill
    \begin{subfigure}[c]{1.0\linewidth}
    \centering
     \includegraphics[width=1\linewidth]{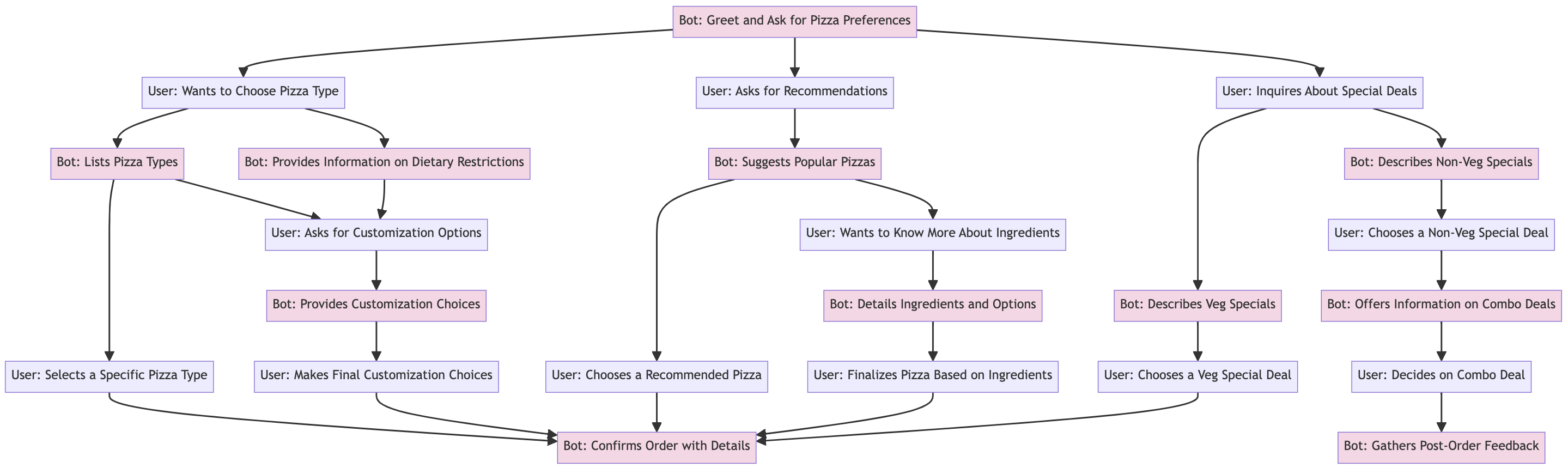}
     \caption{}
\end{subfigure}
\caption{Data-driven (a) and Intrinsic (b) flows for the order pizza domain from MetaLWoz.}
\end{figure*}

\begin{figure*}[h]
      \begin{subfigure}[c]{1.0\linewidth}
    \centering
     \includegraphics[width=1\linewidth]{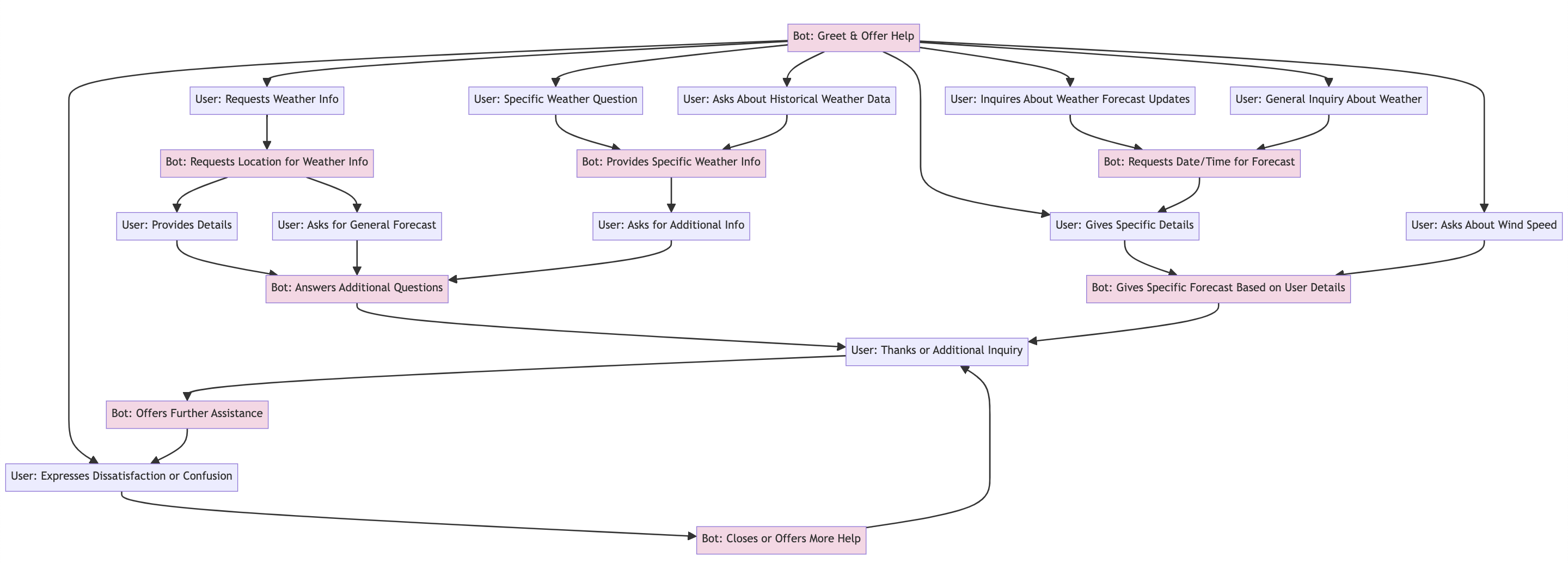}
     \caption{}   
\end{subfigure}
    \hfill
    \begin{subfigure}[c]{1.0\linewidth}
    \centering
     \includegraphics[width=1\linewidth]{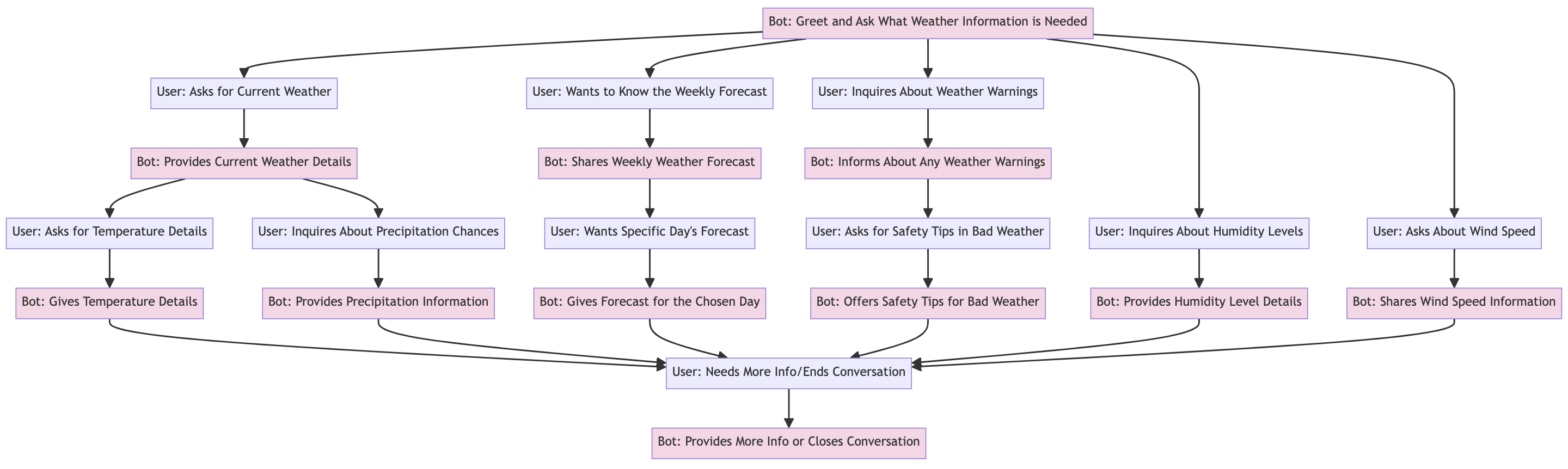}
     \caption{}
\end{subfigure}
\caption{Data-driven (a) and Intrinsic (b) flows for the order weather domain from MetaLWoz.}
\end{figure*}

\begin{figure*}[h]
    \centering
    \includegraphics[width=1\linewidth]{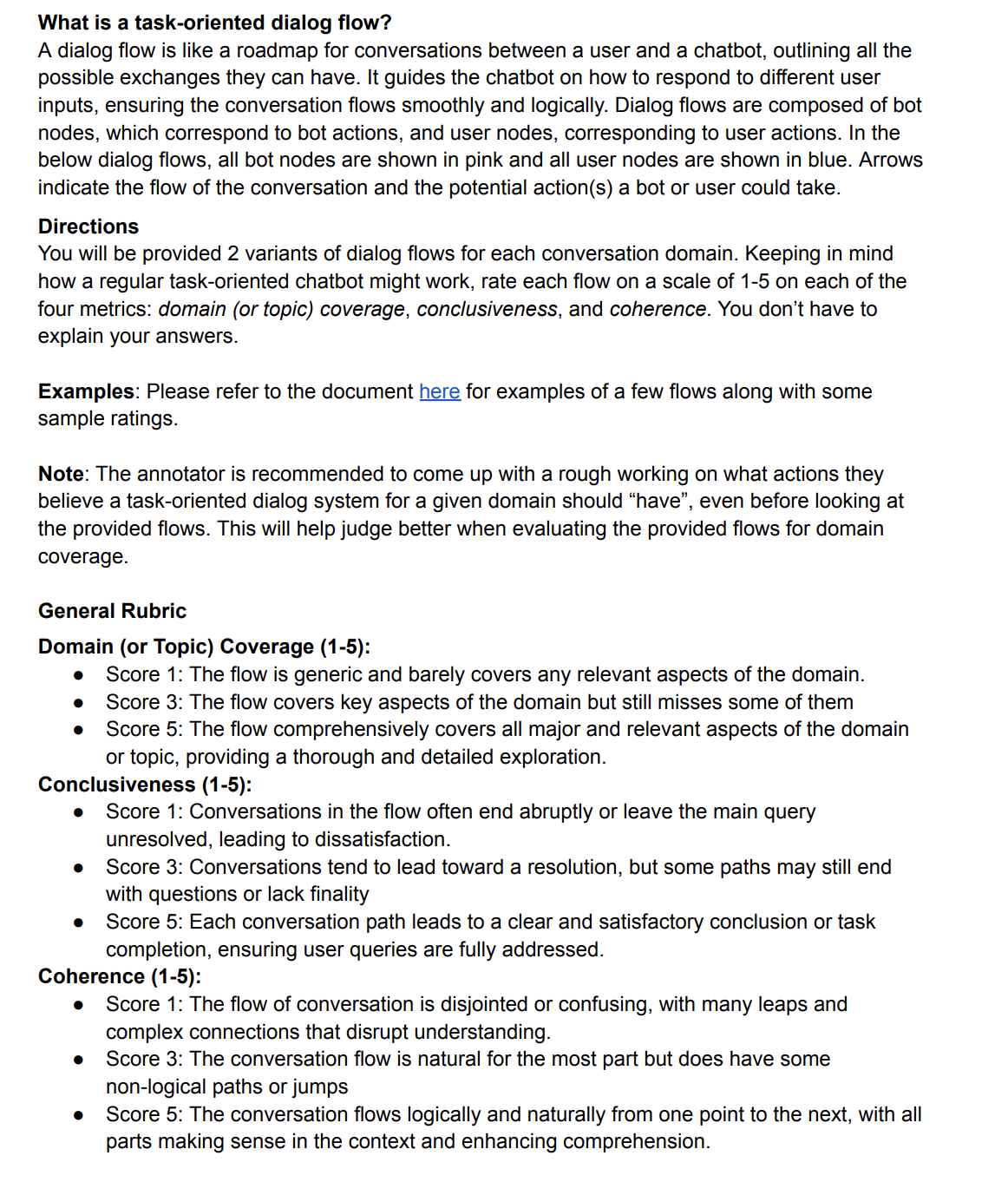}
     \caption{Evaluation Instructions for Human Annotators} 
\end{figure*}    

\end{document}